\pdfoutput=1

\documentclass[11pt]{article}

\usepackage[]{EMNLP2023}

\usepackage{times}
\usepackage{latexsym}

\usepackage[T1]{fontenc}

\usepackage[utf8]{inputenc}

\usepackage{microtype}

\usepackage{inconsolata}

\usepackage{cmap}
\usepackage[T1]{fontenc}
\usepackage{graphics} 
\usepackage{epsfig} 
\usepackage{mathptmx} 
\usepackage{amsmath} 
\usepackage{amssymb}  
\usepackage{comment}
\usepackage{multirow}

\usepackage{float}
\usepackage{booktabs}
\usepackage{graphicx}
\usepackage{url}

\usepackage{comment}
\usepackage{siunitx}
\usepackage{relsize}
\usepackage{ifthen}
\usepackage[colorinlistoftodos]{todonotes}

\usepackage[caption=false]{subfig}

\usepackage[vlined,ruled,linesnumbered]{algorithm2e}
\usepackage{graphics} 
\usepackage{rotating}
\usepackage{color}
\usepackage{enumerate}
\usepackage[T1]{fontenc}
\usepackage{psfrag}
\usepackage{epsfig} 
\usepackage{hyperref}
\usepackage{booktabs}
\usepackage{graphicx,url}
\usepackage{multirow}
\usepackage{array}
\usepackage{latexsym}
\usepackage{amsfonts}
\usepackage{amsmath}
\usepackage{amssymb}
\usepackage{xstring}
\usepackage[noend]{algorithmic}
\usepackage{multirow}
\usepackage{xcolor}
\usepackage{prettyref}
\usepackage{flexisym}
\usepackage{bigdelim}
\usepackage{breqn} 
\usepackage{listings}

\usepackage{enumitem}
\usepackage{xspace}
\usepackage{bm}
\graphicspath{{./figures/}}
\usepackage{tikz}
\usetikzlibrary{matrix,calc}

\usepackage{mdwlist}

\makecompactlist{itemize}{stditemize}

\newrefformat{prob}{Problem\,\ref{#1}}
\newrefformat{def}{Definition\,\ref{#1}}
\newrefformat{sec}{Section\,\ref{#1}}
\newrefformat{sub}{Section\,\ref{#1}}
\newrefformat{prop}{Proposition\,\ref{#1}}
\newrefformat{app}{Appendix\,\ref{#1}}
\newrefformat{alg}{Algorithm\,\ref{#1}}
\newrefformat{cor}{Corollary\,\ref{#1}}
\newrefformat{thm}{Theorem\,\ref{#1}}
\newrefformat{lem}{Lemma\,\ref{#1}}
\newrefformat{fig}{Fig.\,\ref{#1}}
\newrefformat{tab}{Table\,\ref{#1}}

\newcommand{\bdmath}{\begin{dmath}}
\newcommand{\edmath}{\end{dmath}}
\newcommand{\beq}{\begin{equation}}
\newcommand{\eeq}{\end{equation}}
\newcommand{\bdm}{\begin{displaymath}}
\newcommand{\edm}{\end{displaymath}}
\newcommand{\bea}{\begin{eqnarray}}
\newcommand{\eea}{\end{eqnarray}}
\newcommand{\beal}{\beq \begin{array}{ll}}
\newcommand{\eeal}{\end{array} \eeq}
\newcommand{\beas}{\begin{eqnarray*}}
\newcommand{\eeas}{\end{eqnarray*}}
\newcommand{\ba}{\begin{array}}
\newcommand{\ea}{\end{array}}
\newcommand{\bit}{\begin{itemize}}
\newcommand{\eit}{\end{itemize}}
\newcommand{\ben}{\begin{enumerate}}
\newcommand{\een}{\end{enumerate}}

\newcommand{\hide}[1]{}

\newcommand{\hiddenText}{{\color{gray} hidden text.}}
\newcommand{\hideWithText}[1]{\hiddenText}

\DeclareMathOperator*{\argmax}{arg\,max}

\newcommand{\blue}[1]{{\color{blue}#1}}

\newcommand{\linkToPdf}[1]{\href{#1}{\blue{(pdf)}}}
\newcommand{\linkToPpt}[1]{\href{#1}{\blue{(ppt)}}}
\newcommand{\linkToCode}[1]{\href{#1}{\blue{(code)}}}
\newcommand{\linkToWeb}[1]{\href{#1}{\blue{(web)}}}
\newcommand{\linkToVideo}[1]{\href{#1}{\blue{(video)}}}
\newcommand{\linkToMedia}[1]{\href{#1}{\blue{(media)}}}
\newcommand{\award}[1]{\xspace} 

\title{Leveraging Large (Visual) Language Models for \\Robot 3D Scene Understanding}

\author{William Chen \and Siyi Hu \and Rajat Talak \and Luca Carlone \\
  Massachusetts Institute of Technology \\
  \texttt{\{verityw, siyi, talak, lcarlone\}@mit.edu}}

\begin{document}
\maketitle
\begin{abstract}
Abstract semantic 3D scene understanding is a problem of critical importance in robotics. As robots still lack the common-sense knowledge about household objects and locations of an average human, we investigate the use of pre-trained language models to impart common sense for scene understanding. We introduce and compare a wide range of scene classification paradigms that leverage \textit{language only} (zero-shot, embedding-based, and structured-language) or \textit{vision and language} (zero-shot and fine-tuned). We find that the best approaches in both categories yield $\sim 70\%$ room classification accuracy, exceeding the performance of pure-vision and graph classifiers. We also find such methods demonstrate notable generalization and transfer capabilities stemming from their use of language. 
\end{abstract}

\section{Introduction}
3D scene understanding is a key challenge in robotics. For robots to see widespread deployment, they must be able to not only map/localize in many environments, but also have a \textit{semantic understanding} of said environments and the entities within them. If a robot is told to ``fetch a spoon,'' it should infer that spoons are found in kitchens, which are usually characterized by things like ovens and sinks, and then use all this knowledge to navigate to a task-appropriate room.

\begin{figure*}[t]
    \centering
    \includegraphics[width=.8\textwidth]{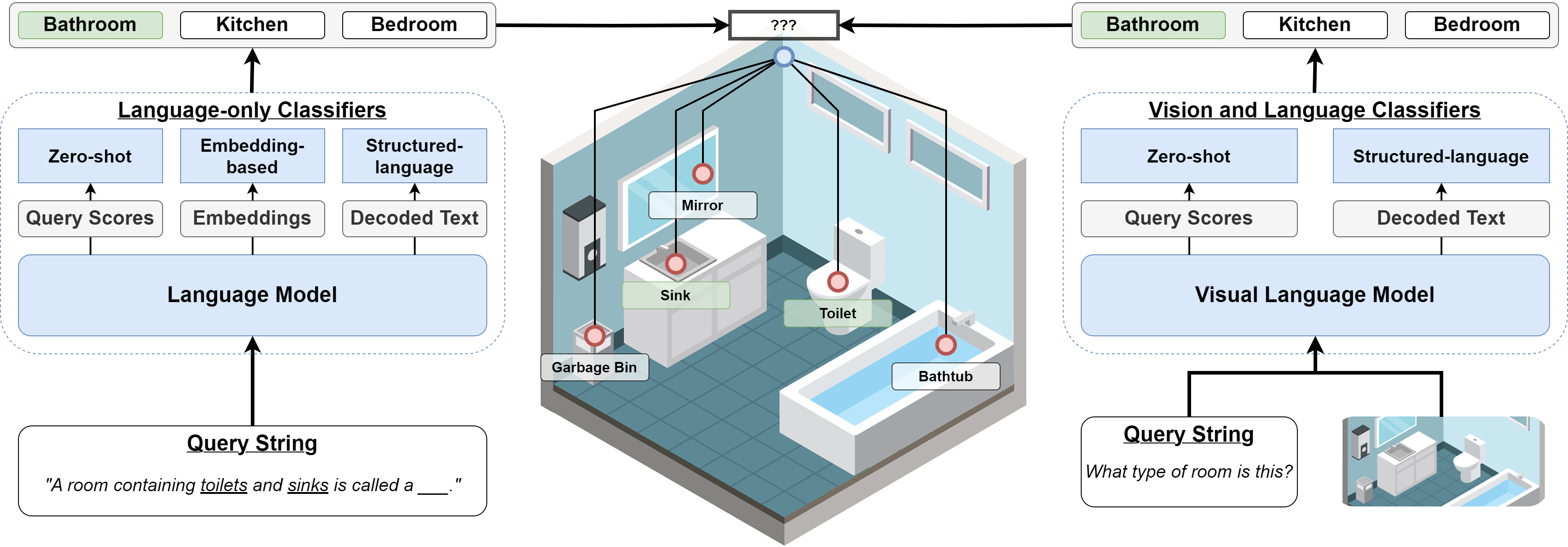}
    \caption{3D scene graph example. We use (V)LMs to attach high-level labels to nodes (e.g., to label rooms) using lower-level information (e.g., contained objects).}
    \label{fig:sgexample}
\end{figure*}

These aspects are typically inferred using metric-semantic simultaneous localization and mapping (SLAM) algorithms, wherein a robotic agent maps its environment, determines its location within it, and annotates the map with semantic information \cite{Bowman17icra}. Modern spatial perception systems, like Kimera \cite{Rosinol21ijrr-Kimera} and Hydra \cite{Hughes22rss-hydra}, arrange this data in 3D scene graphs --  data structures wherein nodes represent locations and entities (e.g., buildings, rooms, and objects), while edges represent spatial relationships (see Fig. \ref{fig:sgexample}).
Nodes can hold geometric information (like pose and bounding box) for entities and places in the scene. However, attaching semantic labels to these nodes still remains a major open obstacle, especially for nodes corresponding to high-level spatial concepts, like rooms and buildings. To label a room node, the system must consider what objects are in the room (e.g., if it contains a stove, the room is likely a kitchen). This necessitates a ``common-sense'' mechanism to provide such knowledge. 

One candidate for imparting this common sense is by using pre-trained language models (LMs). Being trained on large text corpora, they capture some of the semantic information within said datasets -- a LM may learn that the sentence ``Bathrooms contain \underline{\hspace*{.5cm}}.'' is best finished with ``toilets.''. Furthermore, being what \citet{vonHumboldt88-language, Chomsky65-syntax} called an ``infinite use of finite means,'' language allows arbitrary common-sense queries to be compactly made and evaluated, including ones with novel concepts. This is important for spatial perception, as \textit{deployed robots will naturally come across many objects that engineers did not expect during development}. Being able to do inference over novel concepts to suit a given task would thus be highly beneficial.

In this work, we consider ways to leverage pre-trained LMs for abstract robot scene understanding tasks -- specifically, room classification. We show how such models provide useful priors that enable efficient zero-shot or fine-tuned performance. We find that a structured-language approach wherein a LM is fine-tuned for room classification based on strings detailing spatial and semantic object information outperforms similar language-based approaches that only consider object labels (and discards quantitative spatial information, due to it being difficult to express in natural language).

These experiments only reason about language descriptions, discarding visual information. However, such features still provide a useful semantic signal, so we also consider methods for using visual language models (VLMs) for robot scene understanding. We find that visual-linguistic approaches achieve better performance when combining vision with object labels in comparison to vision-only methods (while maintaining minimal hand-engineering effort). We thus argue that combining semantic spatial perception systems with the common-sense priors found in pre-trained (V)LMs enables effective abstract scene understanding.

\section{Related Works}
\label{sec:relatedworks}
\textbf{Metric-Semantic SLAM:} As high-level semantic understanding is vital for human-robot interaction and planning, there has been significant interest in combining classical methods with deep learning for metric-semantic SLAM. Such works generally focus on low-level representations, such as object-centric approaches \cite{Bowman17icra, Dong17cvpr-XVIO, Nicholson18ral-quadricSLAM, Ok21icra-home}, dense approaches \cite{Behley19iccv-semanticKitti, Grinvald19ral-voxbloxpp, Rosu19ijcv-semanticMesh}, or a hybrid of the two \cite{Li16iros-metricSemantic, McCormac183dv-fusion++}. However, these methods generally disregard higher-level semantic labeling which are needed for many planning and reasoning tasks.

\textbf{Hierarchical Mapping:} An alternative approach is to consider hierarchical maps, which represent the robot's environment at different levels of abstraction. Works like \citet{Ruiz-Sarmiento17kbs-multiversalMaps, Galindo05iros-multiHierarchicalMaps} divide robot knowledge into spatial and semantic information, then anchoring the former to the latter. Our work focuses on scene graphs as a hierarchical map representation. Such data structures were first commonly used for visual relation tasks \cite{Lu16eccv-visualRelations, Johnson18cvpr-imageGeneration}, but have since been generalized to 3D, where they have found success in robotics \cite{Armeni19iccv-3DsceneGraphs, Rosinol20rss-dynamicSceneGraphs, Rosinol21ijrr-Kimera, Hughes22rss-hydra}. Nevertheless, anchoring semantics to spatial information in 3D scene graphs remains difficult, motivating us to look towards LMs.

\textbf{Language and Robotics:} Using LMs for robotics has been a rapidly-expanding area of research. Papers like \citet{Tellex11aaai-nlcommands, Sharma22rss-robotPlansLanguage, Lynch20rss-languageconditionedil, Shridhar21corl-cliport, Kollar10hri-directions, Tellex14icra-manipLanguagePlanner, Matuszek13-robotLanguageParser, Ahn22arxiv-sayCan} have mainly leveraged language for communicating goals or instructions to a robot to plan around or execute, often using LMs as intuitive priors over appropriate actions, rewards, or dynamics. For scene understanding, \citet{Huang23icra-vlmaps} use occupancy maps with language embeddings attached, while \citet{Kerr23-lerf, Shafiullah22-clipfields} connect such information to neural radiance fields. These works focus on lower-level semantics, compared to the more abstract semantics considered here. Past approaches for room classification use an explicit Bayesian probabilistic framework for determining room labels based on detected objects \cite{Chaves20appis-roomcat}. However, such methods remain hard to generalize to new rooms and objects. To address these shortcomings, we explore the ability for large LMs to be used as common-sense mechanisms for robot scene understanding.

\section{Language-only Methods}
Modern language models are generally transformers \cite{Vaswani17neurips-transformer} that learn distributions over strings. We write $\Lambda(W) \approx \log p(W)$ to be a LM's estimated log probability score for string $W$. LMs are also typically able to embed texts into semantically meaningful vectors, which can be useful for fine-tuning in down-stream tasks. When pre-trained on large data corpora, all these models' outputs can reflect common world knowledge \cite{Li23icml-LaMPP}. We thus make use of these capabilites for scene understanding.

\subsection{Query Strings and Informativeness}
If we want to use LMs to reason about a room's class, we first need to project its semantic features into language. We start by building query strings describing a given room whose label we wish to infer. For this, we assume access to a list of objects within. This list can be inferred by existing mapping techniques \cite{Rosinol21ijrr-Kimera, Hughes22rss-hydra}. For our experiments, we use ground truth object labels from our considered dataset.

Putting \textit{all} objects in a room into the query may result in poor performance, as the queries may be dominated by uninformative, ubiquitous objects (e.g., lights). We thus draw from Grice's conversational maxims and rational speech acts \cite{Grice75-conversationalMaxims, Goodman16-rsa}. In this framework, a pragmatic speaker chooses an utterance based on how a listener would interpret a literal speaker. E.g., a literal speaker might describe a bedroom as ``containing chairs,'' which is true, but ambiguous to a listener who is trying to discern the room's label. A pragmatic speaker would thus instead say the room contains a \textit{bed}, which is much more informative and disambiguating.

To implement this in our queries, we only include the $k$ objects most informative for room classification, noting that \textit{objects which appear in fewer room types are more informative}, as their presence heavily implies certain rooms. Quantitatively, these objects have more \textit{non}-uniform distributions $p(r \ |\ o )$, where $o \in L_O$ is the object label, $r \in L_R$ is the room label, and $L_{O, R}$ are the sets of all possible objects/rooms respectively. We compute these conditional probabilities in two ways:
\begin{itemize}
    \item \textbf{Using ground-truth co-occurrences}. When using these empirical conditionals, we apply Laplace smoothing. This requires task-specific data.
    \item \textbf{Using proxy co-occurrence probabilities by querying LMs.} Specifically:
    \begin{equation}
        p(r \ |\ o) \approx \frac{\exp \Lambda(W_{o,r})}{\sum_{r' \in L_R} \exp\Lambda(W_{o,r})}
    \end{equation}
    where $W_{o,r}$ is the query string ``A room containing $o$ is called a(n) $r$.''
\end{itemize}
With $p(r \ |\ o)$ available, a natural measure of its non-uniformity (and thus informativeness) is entropy:
\begin{equation}
    H_{o} = -\sum_{r \in L_R} p(r\ |\ o) \log p(r\ |\ o)
\end{equation}
Entropy is maximized when $p(r\ |\ o)$ is uniform and minimized when one-hot, so more informative objects have \textit{lower} $H_{o}$. Thus, to pick objects for the queries, we take the $k$ different lowest-entropy present objects:
\begin{equation}
     O_{\text{best}} = \underset{o \in O}{\text{argmin\_$k$}} \ \left[ H_{o} \right]
\end{equation}
where $O$ is the set of all object labels contained within the considered room. We now have $k$ objects $O_{\text{best}}$ which can be used to infer the room label.

\subsection{Zero-shot Approach}
For the zero-shot approach, we construct $|L_R|$ query strings, one per room label:
\begin{equation}
    \begin{split}
        W_{r} =\ &\text{``A room containing $o_1$, $o_2$, ... and $o_k$}\\
        &\text{is called a(n) $r$.''}\ \forall r \in L_R   
    \end{split}
\end{equation}
where $o_{1...k} \in O_{\text{best}}$ are ordered by ascending entropy. All these queries are scored via LM, with the final estimated room label $\hat{r}$ being whichever one yields the highest query sentence probability:
\begin{equation}
    \hat{r} = \argmax_{r} \Lambda(W_{r})
\end{equation}
We note that this can similarly be done by prompting the LM to \textit{generate} its inferred room type. However, for evaluative purposes, we follow works like \cite{Ahn22arxiv-sayCan} and opt for this scoring approach to constrain the LM to our considered labels.

\subsection{Embedding-based Approach}
For our embedding-based fine-tuning approach, we create a query of the form:
\begin{equation}
        W =\ \text{``This room contains $o_1$, $o_2$, ... and $o_k$.''}
        \label{eq:ftquery}
\end{equation}
This string is then fed into a LM to produce a summary embedding vector. Finally, the embedding is fed into a trained classifier, which produces $|L_R|$ prediction logits corresponding to the room labels, with the inferred room label corresponding to the maximum logit. We choose this network to be a shallow multi-layer perceptron.

\subsection{Structured-data Approach}
The above approaches limit LM inputs to natural language queries. This is restrictive, as robot perception systems tend to detect spatial features, like room size, object poses, and bounding boxes, that are not commonly expressed in natural language. Humans tend to describe spatial relations qualitatively (e.g., ``to the left of'' or ``on top of''), so numeric spatial features may not be processed well via pre-trained LMs. These features are nonetheless useful for room classification: e.g., hallways are often characterized by being \textit{long}.

We thus consider a \textit{structured-data} approach wherein a LM accepts description strings that can easily express spatial features, but are not \textit{natural} language -- e.g., serialized tabular data. This way, the inputs are more expressive, but still capitalize on the LM's common-sense semantic understanding of objects and rooms.

Past approaches for parsing and classifying structured data involve projecting structured data into natural language, e.g., via templating \cite{Hegselmann23-tabllm}. We instead choose to follow \cite{Chen20-fewshotNLG} and fine-tune an encoder-decoder LM to encode structured data strings containing the room's axis-aligned dimensions and each contained object's label, count, and position (relative to the center of the room), then decode probabilities for each possible room label conditioned on the structured description. The highest-scoring one is the inferred label. See Appendix \ref{app:structured_language_query} for details.

\section{Vision-and-language Methods}
In grounded tasks, LMs require observations be projected into string space. Some features are invariably lost in this process. In our case, we only use object labels and spatial features, while visual features like color or texture go unused. Such data is still useful in room classification: e.g., bathrooms often have white tiles that other rooms do not have.

We thus look to visual language models: pretrained models that connect text and language. For instance, \citet{Radford21arxiv-CLIP} uses a contrastive objective to learn a shared embedding space for vision and language, allowing one to match images with text. Other approaches generate text conditioned on input images and text prompts, usually for visual question answering (VQA) or captioning \cite{Li22-blip, Li23-blip2}. We try to use such models to classify locations based on previously-available object information \textit{and} egocentric room images. 

\subsection{Zero-shot Approach}
\label{subsec:vlm_zeroshot}
We consider a VLM equivalent to the language-only zero-shot approach. For contrastive VLMs, we follow the standard zero-shot classification technique \cite{Radford21arxiv-CLIP} and embed both room label strings and images from the considered room. The inferred label is whichever room string best matches the images in embedding space (as determined by cosine distance).

For generative VLMs, we input the images and a prompt (``What type of room is this?'' for VQA models and ``This room is a(n) \textit{r}.'' for captioners). Such prompts can be prepended with object descriptions -- an advantage they have over the contrastive VLMs. The model then follows up the prompt with each possible room label, with the highest-probability label being the inferred one.

\subsection{Fine-tuning Approach}
This approach is the same as the zero-shot VLM approach, but with the model fine-tuned to output room labels. We train a VQA VLM to correctly answer ``What type of room is this?'' when given images of said room. Again, at inference time, the inferred room label is whichever one answers the prompt question with the highest probability.

Note that, for both these approaches, we use captioning/VQA VLMs on a single image at a time. We detail how to get an overall room classification from many images in Sec \ref{subsec:vlm-trial-specs}. Our approach can be easily generalized to VLMs that accept multiple input images, like \citet{Alayrac22neurips-flamingo}. Due to computation limits, we leave such experiments to future work.

\section{Experimental Setup}
\subsection{Datasets}
\label{subsec:dataset}
We evaluate our algorithms on scene graphs produced from the Matterport3D dataset \cite{Chang173dv-Matterport3D}, which is commonly used in robot navigation tasks \cite{Savva19iccv-habitat, Yadav22-habitatchallenge, Szot21neurips-habitat}. This dataset can rapidly render rooms and contained objects, each with labels, pose, and bounding boxes that we use to create scene graphs. Objects are assigned labels from two label spaces: mpcat40 (35 labels) and nyuClass (201 labels) \cite{Silberman12eccv}. In total, there are $\sim 1870$ rooms, each with one of 23 room labels. See Appendix \ref{app:mp3d_scenegraph} for scene graph construction details. We now consider how datasets are constructed for language-only and vision-and-language approaches respectively.

\textbf{Language-only:} We divide the buildings into a 50/20/30 train/validation/test split for each label space. To produce queries for our embedding-based approach, we bootstrap subsets of the most informative objects per room to put into the query. We generate four such datasets by varying object label space (nyuClass/mpcat40) and co-occurrences used for object selection (ground truth/proxy). We use RoBERTa-large as our LM embedder \cite{Liu19iclr-roberta}. See Appendix \ref{app:embedding_bootstrapping} for more details. For the structured-language dataset, for each room, we create a string of the form in Appendix \ref{app:structured_language_query} summarizing the room dimensions and object positions and labels (for a given label space). Unlike the zero-shot/embedding approaches, said strings contain \textit{all} objects in a given room.

Each dataset for a certain label space is produced from the same splits. All approaches and baselines are tested on the same test split. For completeness, approaches that do not require training are also evaluated on the entire dataset.

\textbf{Vision and Language:} For each room, we sample and save images from 100 random freespace camera poses (see Appendix \ref{app:VL_dataset_details}). For methods that require training, we divide the rooms into a 40/20/40 train/validation/test split. Note that (i) all images for a given room belong to the same group and (ii) this is a different divide than for language-only, as it includes rooms with no objects (as visual features can still be extracted from them). 

\subsection{Baselines}
\label{subsec:baselinetrials}
We consider three baselines. First, we use ground-truth co-occurrence data for a \textit{statistical baseline}. We approximate the probability of a room label as the product of the conditional probabilities of the room given each object individually:
\begin{equation}
    p(r \ |\ O) \approx \prod_{o_i \in O} p(r\ |\ o_i)
\end{equation}
where the conditionals $p(r \ |\ o_i)$ are empirically estimated from the dataset. The inferred label is thus $\argmax_{r} p(r\ |\ O)$. Second, we train a \textit{GraphSage graph neural network baseline} \cite{Hamilton17nips-GraphSage} to predict rooms given objects using the language-only dataset splits.

Finally, as a baseline for vision-based methods, we use the vision-and-language dataset splits to train a ResNet-50 \cite{He16cvpr-ResNet} model to predict room label logits given images from the room. We do this from scratch and starting from the pre-trained weights. See Appendix \ref{app:trainingdetails} for details.

\subsection{Language-only Trial Specifications}
\textbf{Zero-shot:} We vary the ground truth/proxy object-room co-occurrences (when computing object entropy) and the object label spaces for four total conditions. We choose $k=3$ objects per room to create the corresponding queries (or all if the room contains fewer than three objects). For all trials, we use GPT-J \cite{Wang21-gpt-j} for both inference and generating proxy co-occurences. Due to hardware limitations, we use the half-precision release of the model.

\textbf{Embedding-based:} We train head networks on each of the four embedding datasets. See Appendix \ref{app:trainingdetails} for more details. We also run two generalization experiments. First, we train the network while holding out all nyuClass rooms whose query strings contain chairs, sinks, toilets, beds, and washing machines, then we test on these held out datapoints. This is done with ground-truth co-occurrences. Second, we train models on mpcat40 data while \textit{testing} on nyuClass data in order to see if they can accommodate and generalize to a different, larger input label space. In this case, we divide the mpcat40 dataset using a 40/60 training/validation split, use the entire nyuClass dataset for testing, and vary the co-occurrence type (ground truth and proxy).

\textbf{Structured-language:} We fine-tune a pre-trained T5 LM to encode structured-language strings and decode (score) room labels \cite{Raffel20jmlr-t5}. We train for 5 epochs and test on the weights with highest validation accuracy. See Appendix \ref{app:trainingdetails} for more details. We also run some ablation trials, removing the room bounding box dimensions, object positions, or both.

\subsection{Vision-and-language Trial Specifications}
\label{subsec:vlm-trial-specs}
\textbf{Zero-shot:} We test CLIP, BLIP, and BLIP-2's zero-shot visual room classification abilities \cite{Radford21arxiv-CLIP, Li22-blip, Li23-blip2}. Of these models, CLIP is contrastive while the BLIP models are generative, so we adopt the VQA and captioning approaches detailed in Sec. \ref{subsec:vlm_zeroshot} respectively.

As each room has many images, we measure both the portion of individual images and overall rooms classified correctly as image-wise and room-wise accuracies, respectively. A room is classified correctly if, when the scores for each label are summed over all images from said room, the correct label has the highest total score. This is \textit{different} than a plurality vote wherein the most-predicted room label (out of all image predictions) is the room label, as our method incorporates model uncertainties by using the output logits/scores.

\textbf{Fine-tuning:} We fine-tune the BLIP VQA VLM to decode room labels based on images and descriptions. The setup is the same as the zero-shot case, but with the introduction of fine-tuning. See Appendix \ref{app:trainingdetails} for more details. Again, we measure image- and room-wise accuracies with the same maximum score metric as above.

\subsection{Real Scene Graph Trial Specification}
To highlight how our algorithms fit together with modern robot spatial perception software, we run our approaches on a real scene graph. We use Hydra \cite{Hughes22rss-hydra} to generate a scene graph of the apartment environment in the uHumans2 \cite{Rosinol21ijrr-Kimera} photorealistic simulator. The scene graph is queried for the objects in the same room as the robot at each frame. We then run the fine-tuned vision-and-language algorithm from Section \ref{subsec:vlm-trial-specs} on the current image to produce a distribution over room classes based on the objects from the scene graph and the agent's current observation. As the scene graph also yields the current room of the agent, we aggregate and track room-wise label statistics as well. See Appendix \ref{app:real-scene-graph-details} for more details.

\section{Results}
\begin{table*}[ht]
\centering
\resizebox{\textwidth}{!}{%
\begin{tabular}{@{}ccccccccccc@{}}
\toprule
\multirow{3}{*}{} & \multicolumn{2}{c}{\multirow{2}{*}{\textbf{Baselines}}} & \multicolumn{2}{c}{\multirow{2}{*}{\textbf{Zero-shot}}} & \multicolumn{2}{c}{\textbf{Embedding-based}} & \multicolumn{4}{c}{\textbf{Structured-language}} \\ \cmidrule(l){6-11} 
 & \multicolumn{2}{c}{} & \multicolumn{2}{c}{} & \multicolumn{2}{c}{\textbf{RoBERTa}} & \multicolumn{2}{c}{\textbf{With Position}} & \multicolumn{2}{c}{\textbf{No Position}} \\ \cmidrule(l){2-11} 
 & \textbf{Statistical} & \textbf{GraphSage} & \textbf{GT} & \textbf{Proxy} & \textbf{GT} & \textbf{Proxy} & \textbf{GT} & \textbf{Proxy} & \textbf{GT} & \textbf{Proxy} \\ \midrule
\textbf{nyuClass} & 57.7\% (50.6\%) & 64.2\% & 52.2\% (53.6\%) & 27.2\% (28.2\%) & 65.5\% & 57.6\% & \textbf{69.1\%} & 68.7\% & 67.3\% & 67.7\% \\
\textbf{mpcat40} & 44.7\% (46.9\%) & 59.1\% & 48.9\% (50.1\%) & 27.5\% (27.3\%) & 63.9\% & 58.5\% & \textbf{68.3\%} & 67.7\% & 63.9\% & 63.7\% \\ \bottomrule
\end{tabular}%
}
\caption{Test-split accuracies for all language-only approaches. Methods that do not require training have full dataset accuracy reported in parentheses. Highest test split accuracies are bolded.}
\label{tab:language-testaccs}
\end{table*}
\begin{table*}[ht]
\centering
\resizebox{\textwidth}{!}{%
\begin{tabular}{@{}ccccccccccccc@{}}
\toprule
\multirow{3}{*}{} & \multicolumn{7}{c}{\textbf{Zero-shot}} & \multicolumn{2}{c}{\textbf{Baseline}} & \multicolumn{3}{c}{\textbf{Fine-tuning}} \\ \cmidrule(l){2-13} 
 & \textbf{CLIP} & \multicolumn{3}{c}{\textbf{BLIP VQA}} & \multicolumn{3}{c}{\textbf{BLIP-2 Captioner}} & \multicolumn{2}{c}{\textbf{ResNet-50}} & \multicolumn{3}{c}{\textbf{BLIP VQA}} \\ \cmidrule(l){2-13} 
 & \textbf{None} & \textbf{None} & \textbf{nyuClass} & \textbf{mpcat40} & \textbf{None} & \textbf{nyuClass} & \textbf{mpcat40} & \textbf{\begin{tabular}[c]{@{}c@{}}From \\ Pretrained\end{tabular}} & \textbf{\begin{tabular}[c]{@{}c@{}}From \\ Scratch\end{tabular}} & \textbf{None} & \textbf{nyuClass} & \textbf{mpcat40} \\ \midrule
\textbf{\begin{tabular}[c]{@{}c@{}}Room-\\ wise\end{tabular}} & \begin{tabular}[c]{@{}c@{}}36.5\%\\ (32.7\%)\end{tabular} & \begin{tabular}[c]{@{}c@{}}37.8\%\\ (36.2\%)\end{tabular} & \begin{tabular}[c]{@{}c@{}}37.0\%\\ (36.6\%)\end{tabular} & \begin{tabular}[c]{@{}c@{}}37.1\%\\ (37.5\%)\end{tabular} & \begin{tabular}[c]{@{}c@{}}47.5\%\\ (45.7\%)\end{tabular} & \begin{tabular}[c]{@{}c@{}}47.9\%\\ (45.7\%)\end{tabular} & \begin{tabular}[c]{@{}c@{}}48.0\%\\ (46.2\%)\end{tabular} & 51.1\% & 26.2\% & 53.2\% & \textbf{68.6\%} & 65.3\% \\
\textbf{\begin{tabular}[c]{@{}c@{}}Image-\\ wise\end{tabular}} & \begin{tabular}[c]{@{}c@{}}26.5\%\\ (25.9\%)\end{tabular} & \begin{tabular}[c]{@{}c@{}}30.1\%\\ (28.8\%)\end{tabular} & \begin{tabular}[c]{@{}c@{}}35.6\%\\ (34.9\%)\end{tabular} & \begin{tabular}[c]{@{}c@{}}34.4\%\\ (34.3\%)\end{tabular} & \begin{tabular}[c]{@{}c@{}}40.1\%\\ (39.3\%)\end{tabular} & \begin{tabular}[c]{@{}c@{}}45.0\%\\ (43.1\%)\end{tabular} & \begin{tabular}[c]{@{}c@{}}44.5\%\\ (43.0\%)\end{tabular} & 36.8\% & 22.6\% & 47.0\% & \textbf{67.9\%} & 64.1\% \\ \bottomrule
\end{tabular}%
}
\caption{Test-split room- and image-wise accuracies for all language and vision approaches. Methods that do not require training have full dataset accuracy reported in parentheses. Highest test split accuracies are bolded.}
\label{tab:vlm-accuracy}
\end{table*}

We present all results for language-only and vision-and-language (and associated baselines) in Tables \ref{tab:language-testaccs} and \ref{tab:vlm-accuracy} respectively.

\subsection{Language-only Results}
\textbf{Zero-shot:} The zero-shot trials yield room classification accuracies of $27.3 - 52.6\%$ when run on the entire dataset. The ground-truth co-occurrence trials perform better than the statistical baseline evaluated on the whole dataset, which also uses ground truth frequencies. No trial outperforms the GraphSage baseline, but said baseline requires training and cannot be easily extended to additional labels, unlike our approach; by virtue of being zero-shot, the only step needed to adopt the large label space was to compute the informativeness metric for the new objects.
\begin{figure}[t]
    \centering
    \includegraphics[width=.45\textwidth]{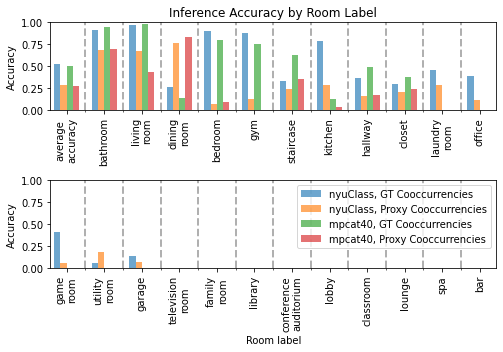}
    \caption{Zero-shot accuracies on all data for all conditions, by room label.}
    \label{fig:cataccs}
\end{figure}
This approach also achieves high accuracies for several common household rooms (bathroom, bedroom, kitchen, and living room). For the best performing trial, accuracies for these key rooms range from $79.2 - 97.1\%$ (see Fig. \ref{fig:cataccs}). There also are two trends for when a room will \textit{not} be classified correctly:
\begin{itemize}
    \item \textbf{Room lacks disambiguating objects:} Bathrooms and bedrooms have objects almost exclusive to them (e.g., toilets and beds), but rooms like lobbies and family rooms only contain more ubiquitous ones (e.g., tables and chairs), and so are harder to identify. 
    \item \textbf{Room is not ``standard'':} Bars, libraries, and spas all more commonly refer to \textit{buildings}, not rooms. We note this could likely be fixed with further prompt tuning and few-shot examples.
\end{itemize}
For the other rooms, the LM shows the desired common sense when classifying them. Our approach also demonstrates generalization, handling the smaller, 35-object label space (mpcat40) \textit{and} the much larger, 201-object label space (nyuClass). In fact, the nyuClass trials result in higher accuracies than their mpcat40 counterparts, as nyuClass's labels are more specific and informative. This benefit is best shown in the following cases:
\begin{itemize}
    \item \textbf{Kitchens \& laundry rooms:} Both rooms are characterized by appliances. While nyuClass provides fine-grained labels (e.g., washing machine vs. stoves), mpcat40 groups all those objects under the broad and ambiguous category of ``appliances,'' making differentiation of the two room labels difficult.
    \item \textbf{Game rooms \& garages:} Game rooms are characterized by recreational objects, like ping-pong/foosball tables. Both appear in nyuClass, but are just ``tables'' in mpcat40, making these rooms easier to identify with the former. Likewise, garage doors (nyuClass) are just called ``doors'' in mpcat40.
\end{itemize}
Finally, all trials take $\sim 1.2$ seconds to infer a room's label.

\textbf{Embedding-based:} The embedding-based approach gets $57.6 - 65.4\%$ test accuracy, beating the statistical baseline ($46.9 - 52.7\%$) and zero-shot approach ($27.2 - 52.2\%$) in all conditions. Unlike for the zero-shot case, this is even true when using proxy co-occurrences, so the model does not explicitly need ground-truth co-occurrences for picking out objects to achieve high performance (though some is learned in training). Trials using ground-truth co-occurrences yield slightly higher accuracies than corresponding GraphSage baselines, while the proxy trials are still competitive. 

Finally, RoBERTa has $20\times$ fewer parameters than GPT-J (used for zero-shot), needing less compute, memory, and time to generate embeddings. However, it is inappropriate for zero-shot evaluation, only yielding $44.5\%$ accuracy on the best condition (nyuClass, ground-truth co-occurrences). Its zero-shot inference speed is also slower: being a MLM, it uses \textit{pseudo}-log likelihood scores \cite{Salazar20acl-PLL}, which increases inference time by $2.5 \times$ for our task (GPT-J takes $\sim40$ min while RoBERTa takes $\sim100$ for the same dataset). Our embedding-based approach thus has better inference speed and performance while still using a smaller LM, at the cost of needing training data.

\begin{table}[ht]
\centering
\resizebox{.75\columnwidth}{!}{%
\begin{tabular}{@{}clclclclcl@{}}
\toprule
\multicolumn{10}{c}{\textbf{Holdout Trial Room Accuracies}} \\
\multicolumn{2}{c}{\textbf{\begin{tabular}[c]{@{}c@{}}Washing\\ Machine\end{tabular}}} & \multicolumn{2}{c}{\textbf{Chair}} & \multicolumn{2}{c}{\textbf{Sink}} & \multicolumn{2}{c}{\textbf{Toilet}} & \multicolumn{2}{c}{\textbf{Bed}} \\ \midrule
\multicolumn{2}{c}{14.6\%} & \multicolumn{2}{c}{40.0\%} & \multicolumn{2}{c}{64.2\%} & \multicolumn{2}{c}{77.6\%} & \multicolumn{2}{c}{83.5\%} \\ \bottomrule
\end{tabular}%
}
\caption{Test accuracies for rooms containing each holdout object.}
\label{tab:emb-gen-accs}
\end{table}

\textbf{Embedding-based Transfer:} Using LMs to embed room descriptions enables some generalization to novel object classes. For the holdout object trials, the model correctly classifies rooms whose queries contain sinks, toilets, and beds $64.2 - 83.5\%$ of the time, even when held out in training (see Table \ref{tab:emb-gen-accs}). The network likely learns to extract essential information on room labels from query embeddings containing only non-held out objects that generalizes to the held out ones. E.g., while toilets are held out, related objects like bathtubs are not. The embeddings for observed queries (``This room contains bathtubs.'') may be similar to that of unobserved ones (``This room contains toilets.''), so the network classifies the latter correctly too. As there are no objects related to washing machines in nyuClass, rooms containing them have comparatively low accuracy. Still, as most rooms have several characteristic objects, holding out a subset of them (or, equivalently, introducing more of them) is generally not an issue.

Regardless, our approach shows promising generalization and transferability. When trained on mpcat40 and tested on nyuClass, it yields $47.1\%$ and $56.6\%$ accuracy for proxy and ground-truth co-occurrences respectively, comparable to the best accuracy for the zero-shot approach ($52.57\%$) and some non-transfer fine-tuning conditions.

\textbf{Structured Language:} We expect the structured-language approach to do at least as well as the embedding approach, as the structured language string's information is a superset of that of the embedding-based inputs. Sure enough, all structured-language approaches have test accuracies from $63.7-69.1\%$, being comparable to or better than all other approaches. We note that ablating room bounding box information does not degrade performance much, but removing object positions is more impactful (especially for mpcat40 trials). Nevertheless, all changes from ablations are relatively small; generally, \textit{structured language of any type seems to be the best overall language-only approach for room classification}.

\subsection{Vision-and-language Results}
\textbf{Zero-shot:} The BLIP-2 trials have the highest zero-shot room-/image-wise accuracies, comparable to that of the fine-tuned ResNet-50 baseline, despite having no task-specific training. Adding object labels (from either label space) to the prompt improves image-wise accuracy, likely as it aids in the classification of uninformative images (e.g., ones facing walls) by providing non-visible room-wise information on contained objects. However, it surprisingly does \textit{not} improve room-wise accuracy, perhaps because the improved image classifications are already for rooms that are visually distinct and thus can be classified from just the good images.

We find that the BLIP-2 approach has slightly lower accuracy than the best language-only zero-shot trial ($45.7 - 46.2\%$ vs. $52.6\%$), though a bit higher than the second-best trial. However, the former needs less prompt engineering and no ground-truth co-occurrence data (for picking objects for the query template in the language-only case), so is easier to implement and use.

\textbf{Fine-tuning:} The fine-tuning approach yields similar accuracies to the structured-language approach. Notably, in contrast to the zero-shot equivalents, the fine-tuned models' image- \textit{and room-wise} accuracies benefit from object labels in the prompt. This suggests that, unlike with zero-shot, the fine-tuned models both use the object labels to correctly classify aforementioned uninformative images \textit{specifically for rooms that would otherwise not be correctly predicted} -- it is not just correcting uninformative images in rooms whose other images are easily correctly classifiable.

Moreover, we see that the fine-tuned trial that does not have object labels achieves a room-wise accuracy of $53.2\%$, compared to the pre-trained ResNet-50 vision-only baseline accuracy of $51.1\%$. These values are similar and both higher than the from-scratch ResNet-50 baseline trial, suggesting that pre-training (be it visual or visual-linguistic) aids in transfer to the room classification domain. 

However, both pre-trained ResNet-50 and the fine-tuned no-label trials fail to beat \textit{any} of the approaches that use fine-tuning in conjunction with object labels (including the language-only approaches). This suggests that, even when trained on domain-specific data, it is difficult for vision alone to classify rooms from arbitrary images within. 

It is thus appropriate to supplement such inferences with room-wise object information, especially if such data is already available from a spatial perception system. In this case, we find that (V)LMs can easily incorporate object labels into their room label inferences, leveraging their common-sense understanding of object-room relations to improve room classification accuracy.

\subsection{Real Scene Graph Results}
For the real scene graph trial, the fine-tuned VLM's room-wise classifications are nearly all correct. The only incorrect room is the dining room, where the VLM assigns similar scores to ``kitchen'' and ``dining room'' (though the former is slightly higher). For all other rooms, the model assigns high scores to the correct label. We present a visualization of image-wise room classifications in \url{https://tinyurl.com/2wj7aan3}. We also give an example visualization image, images from each room, and full room-wise classification scores in Figs. \ref{fig:room_examples}, \ref{fig:example_pred}, and Table \ref{tab:real_dsg_roomwise_logits} respectively, all in Appendix \ref{app:real_scene_graph_results}.

\section{Conclusion}
We show the applicability of large pre-trained LMs and VLMs to the problem of abstract robot scene understanding, particularly in the domain of room classification. We explore an array of language-only and vision-and-language approaches, comparing them with standard statistical and learned baselines. We find that using LMs yields higher overall performance while also having good generalization to held-out objects and transferability to new label spaces. Our results show that these paradigms are promising avenues of development for scalable, sample-efficient, and generalizable robot spatial perception systems.

\section*{Limitations}
Our primary limitation is compute. We ran all LM inference experiments on a single RTX 3080 GPU, which vastly constrains the size of model (and thus the resulting experiments) we could use. 

In particular, while not quantitatively reported, we find that the zero-shot approach empirically benefits greatly from using larger models trained on more data; smaller equivalent models like GPT-Neo \cite{Black21-gpt-neo} or GPT-2 \cite{Radford19-GPT2} yield very poor zero-shot results. We found that the six billion parameter GPT-J was the largest such model we could reliably load, but still had to use the half-precision release due to memory constraints. 

Likewise, our structured language approach only includes room dimensions, object labels, and object positions; some spatial/geometric information, like object pose, is still lost, as they do not fit in our smaller models' token counts (or, even if they do, would take up too much memory).

We thus detail some additional research directions that would be possible with larger compute budgets. We suspect that newer, larger models (on the scale of $10-100$ billion parameters) would improve performance even more, especially when combined with recent advances in instruction and human preference-based tuning. For the VLM experiments, newer models explictly trained for embodied visual-linguistic reasoning \cite{Driess23-palme} would also likely perform very well for our scene understanding tasks. Finally, many common-sense visual understanding problems have shown to work well with Socratic model \cite{Zeng22arxiv-socraticmodels} setups, wherein multiple models (typically covering multiple modalities) interface with each other in natural language.

However, these constraints \textit{do} also reflect a real challenge if such systems are deployed on real mobile robots -- such autonomous systems generally do not have the hardware required to load the latest and largest LMs on them locally. It is therefore promising that, even with somewhat older, smaller models (which \textit{could} feasibly be deployed on a robotic platform), we still achieve relatively good performance in our considered tasks.

It thus would be exciting to see both how one could get the most out of the limited compute of a mobile robot \textit{and} how our scene understanding approaches could be made more effective when the limitation is relaxed.

\section*{Ethics Statement}
When applying large pre-trained models to robotics, it is very important to recognize how the risks associated with such models can be amplified when given embodiment \cite{Bommasani22-foundationModels}. We focus on the relatively innocuous use case of scene understanding/location classification, but even still, the biases captured within our considered models may be reflected in their performance on such tasks. For instance, we evaluate on the Matterport3D dataset, which contains scans of domestic environments. It is safe to say this data represents the homes of a narrow intersection of wealth, class, and culture. It is possible that homes of certain demographics may not conform to the prototypical conceptualization of such locations present within our (V)LMs, \textit{and} that our evaluative metrics do not detect such biases (again, as our testing data is not diverse enough).

Thus, we believe it is important for such data to be diverse and transparently/ethically sourced (while maintaining proper privacy), especially when used to train general-purpose household robots. In addition, we encourage research into more advanced scene understanding techniques that allow for interpretability of inferences and adaptability to a wider range of environments. 


\bibliography{new,myRefs,main}
\bibliographystyle{acl_natbib}

\appendix

\section{Converting Matterport3D to Scene Graphs}
\label{app:mp3d_scenegraph}
To convert semantic meshes from Matterport3D into scene graphs, we create a node for each region and object \cite{Chang173dv-Matterport3D}. Then, we connect all object nodes assigned to a region to that region's room node. We also filter out some regions. While Matterport3D contains outdoor regions as well (``yard,'' ``balcony,'' and ``porch''), we do not perform inference over them, since they are not true rooms and thus would require an alternate query string structure. In addition to outdoor regions, we also remove all rooms with no objects within or with the label ``none.''

Each object is assigned labels from several label spaces. We consider the original labels used by Matterport3D (mpcat40) and the labels used by NYU (nyuClass) \cite{Silberman12eccv}. For both, we filter out nodes belonging to the mpcat40 categories ``ceiling,'' ``wall,'' ``floor,'' ``miscellaneous,'' ``object,'' and any other unlabeled nodes. We remove these categories because they are either not objects within the room or they are ambiguous to the point of being semantically uninformative. However, for nyuClass, we do \textit{not} reject objects classified by mpcat40 as ``object,'' since nyuClass has many more fine-grained and semantically-rich categories which all are mapped to this category. After pre-processing the label spaces in this way, mpcat40 has 35 object labels and nyuClass has 201. Both datasets share a room label space with 23 labels. See Table \ref{tab:labelfreqs} for a breakdown of room label frequencies.

\begin{table}[ht]
\centering
\resizebox{\columnwidth}{!}{%
\begin{tabular}{@{}ccccccc@{}}
\toprule
\textbf{Room Label} &
  \textit{Bar} &
  \textit{Bathroom} &
  \textit{Bedroom} &
  \textit{Classroom} &
  \textit{Closet} &
  \textit{\begin{tabular}[c]{@{}c@{}}Conference \\ Auditorium\end{tabular}} \\ \midrule
\textbf{Occurrences} &
  3 &
  365 &
  251 &
  2 &
  99 &
  16 \\
\textbf{Percentage} &
  0.16\% &
  19.43\% &
  13.37\% &
  0.11\% &
  5.27\% &
  0.85\% \\ \midrule
\textbf{Room Label} &
  \textit{\begin{tabular}[c]{@{}c@{}}Dining \\ Room\end{tabular}} &
  \textit{\begin{tabular}[c]{@{}c@{}}Family \\ Room\end{tabular}} &
  \textit{Game Room} &
  \textit{Garage} &
  \textit{Gym} &
  \textit{Hallway} \\ \midrule
\textbf{Occurrences} &
  74 &
  61 &
  17 &
  14 &
  16 &
  326 \\
\textbf{Percentage} &
  3.94\% &
  3.25\% &
  0.91\% &
  0.75\% &
  0.85\% &
  17.36\% \\ \midrule
\textbf{Room Label} &
  \textit{Kitchen} &
  \textit{\begin{tabular}[c]{@{}c@{}}Laundry \\ Room\end{tabular}} &
  \textit{Library} &
  \textit{\begin{tabular}[c]{@{}c@{}}Living \\ Room\end{tabular}} &
  \textit{Lobby} &
  \textit{Lounge} \\ \midrule
\textbf{Occurrences} &
  78 &
  35 &
  1 &
  71 &
  62 &
  64 \\
\textbf{Percentage} &
  4.15\% &
  1.86\% &
  0.05\% &
  3.78\% &
  3.30\% &
  3.41\% \\ \midrule
\textbf{Room Label} &
  \textit{Office} &
  \textit{Spa} &
  \textit{Staircase} &
  \textit{\begin{tabular}[c]{@{}c@{}}Television \\ Room\end{tabular}} &
  \textit{\begin{tabular}[c]{@{}c@{}}Utility \\ Room\end{tabular}} &
  \textit{\textbf{Total}} \\ \midrule
\textbf{Occurrences} &
  98 &
  44 &
  152 &
  13 &
  16 &
  \textbf{1878} \\
\textbf{Percentage} &
  5.22\% &
  2.34\% &
  8.09\% &
  0.69\% &
  0.85\% &
  \textbf{100\%} \\ \bottomrule
\end{tabular}%
}
\caption{Room label frequencies in pre-processed Matterport3D dataset.}
\label{tab:labelfreqs}
\end{table}

We perform a few additional dataset pre-processing steps to produce the final scene graph dataset with 1878 rooms. First, since some objects are assigned an incorrect region (e.g., toilets are assigned to living rooms, despite (i) that being non-sensible and (ii) the toilet not being within the bounding box of the living room), we check to see if each object is within the bounding box of its assigned region. If not, then it is re-assigned to whichever region's bounding box contains it, and the corresponding scene-graph connection is also made. Second, nyuClass has some misspelled labels (e.g., ``refridgerator'' instead of ``refrigerator''), so we correct all of those too. Lastly, sometimes, a single nyuClass label may be erroneously assigned to multiple mpcat40 labels. This is most problematic when one of the mpcat40 labels is rejected and the other is not. To address this, we use the first mpcat40 label for each nyuClass label that is \textit{not} rejected (e.g., nyuClass label ``stairs'' is mapped to mpcat40 ``miscellaneous,'' which is rejected, and ``stairs,'' which is not, so we keep the latter). However, this means some labels which \textit{should} be rejected are not rejected, so we also manually filter out all nyuClass object labels that are the \textit{same} as those of rejected mpcat40 labels: ``ceiling,'' ``floor,'' and ``wall''.

\section{Embedding-based Bootstrapping Method}
\label{app:embedding_bootstrapping}
To generate training data for the embedding-based method, we take the $n$ most informative objects in each room and find all $k$-object permutations, producing $P^n_k$ query datapoints per room of the form in Eq. \ref{eq:ftquery}, all of which correspond to the room's label. We do this for $(k, n) \in \{ (1,2), (2,3), (3,4)\}$. Models trained on this data will thus be \textit{invariant to object order and number} in the query, and can also handle less informative object labels.

\section{Structured-language Query String}
\label{app:structured_language_query}
For the structured-language approach, we describe a given room using the following string template:
\begin{quote}
Room Size:\\
x \textit{[x room bounding box length]}\\
y \textit{[y room bounding box length]}\\
z \textit{[z room bounding box length]}

Object Locations:\\
\textit{[object 1 label]}\\
x \textit{[x position relative to room center]}\\
y \textit{[y position relative to room center]}\\
z \textit{[z position relative to room center]}

\textit{[object 2 label]}\\
x \textit{[x position relative to room center]}\\
y \textit{[y position relative to room center]}\\
z \textit{[z position relative to room center]}

(repeat for all other objects in the room)
\end{quote}
To fit hardware memory constraints, we omit rooms with over $100$ objects and round all values to three decimal places. Note that a given room might have different objects depending on which object label space is considered, as nyuClass includes labels that would be just classified as ``object'' by mpcat40 (which we reject).

\section{Vision-and-language Dataset Generation}
\label{app:VL_dataset_details}
To construct a dataset of room images for our vision-and-language methods, we use Matterport3D's Pathfinder functionality to generate a top-down occupancy grid map of freespace for a considered height (which we set to the height of a considered room). A single map pixel's length is set to $0.1$ meters. We find all pixels that fall within a given room's bounding box, masking out the rest. To avoid sampling points that are too close to walls and objects, we dilate occupied cells with a $3 \times 3$ convolutional filter. Then, we uniformly sample (without replacement) $100$ unoccupied cell positions (or as many as possible, if the room has fewer than $100$ open cells). Each one is assigned a yaw uniformly at random. Finally, we set the camera to each of those poses, saving the viewed images and corresponding room label.

\section{Training Details}
\label{app:trainingdetails}
For the \textit{GraphSage baseline}, we train for 500 epochs with a learning rate of $5e-3$, weight decay of $1e-4$, hidden state dimension of $16$, dropout of $0.2$, and $2$ iterations of message-passing. 

For the \textit{embedding-based approach}, we train each network for 200 epochs with a batch size of $512$ using cross entropy loss via the Adam optimizer with a learning rate of $1e-4$, $\beta_1, \beta_2 = 0.9, 0.999$, weight decay of $1e-3$, and a StepLR scheduler with step size of $10$ and $\gamma=0.5$ \cite{Kingma17iclr-adam}.

For the \textit{structured-language approach}, we train the T5 LM for 5 epochs with a batch size of 2 (due to memory constraints) and the AdamW optimizer with learning rate of $1e-4$, smoothing term $\epsilon=1e-8$, and no weight decay \cite{Loshchilov19iclr-adamW}.

For the \textit{ResNet baseline}, we train for 10 epochs using the SGD optimizer with a learning rate of $1e-3$, a batch size of $128$, and momentum of $0.9$. We load the checkpoint weights with the highest validation image-wise accuracy.

For the \textit{fine-tuned VLM approach}, we train a BLIP-2 model for 5 epochs with the AdamW optimizer with a learning rate of $5e-6$, batch size of $8$ images (due to memory constraints), weight decay of $0.01$, and smoothing term of $1e-8$.

In trials training with the vision-and-language dataset, we pick $16$ random images per room (as many as possible if there are fewer than $16$). Additionally, for both validation and testing, we take every fourth image. Both these choices are to reduce the training and evaluation time -- more images can be used for a more complete evaluation. 

\section{Real Scene Graph Trial Details}
\label{app:real-scene-graph-details}
For the real scene graph trial, we consider a scene graph built via Hydra using a trajectory in the uHumans2 simulator's apartment environment \cite{Rosinol21ijrr-Kimera}. The trajectory contains egocentric RBG observations at $30$ Hz and odometry data used to construct the scene graph. The scene graph itself contains nodes representing agent positions along the trajectory. We associate each observation with the temporally-closest agent node, thereby assigning each observation a room (and associated contained objects).

As the observations are in a different domain than Matterport3D, we fine-tune the BLIP VQA VLM on the entire vision-and-language train \textit{and} validation datasets, then evaluate on the test set. For ease of visualization, we limit ourselves to the following labels in training, validation, and deployment: bathroom, bedroom, dining room, family room, kitchen, living room, lounge, office, and television room. We use the same hyperparameters as presented in Appendix \ref{app:trainingdetails}.

Finally, for each image, we generate a descriptive query question string containing the room's contained objects (from the scene graph) and feed both into the fine-tuned VLM for prediction. We show the image-wise prediction distribution in the associated visualization video, but also compute room-wise distribution statistics (based on the room that the scene graph says the agent is in at each timestep). 



\section{Real Scene Graph Results}
\label{app:real_scene_graph_results}

\begin{figure}[h]
    \centering
    \includegraphics[width=\columnwidth]{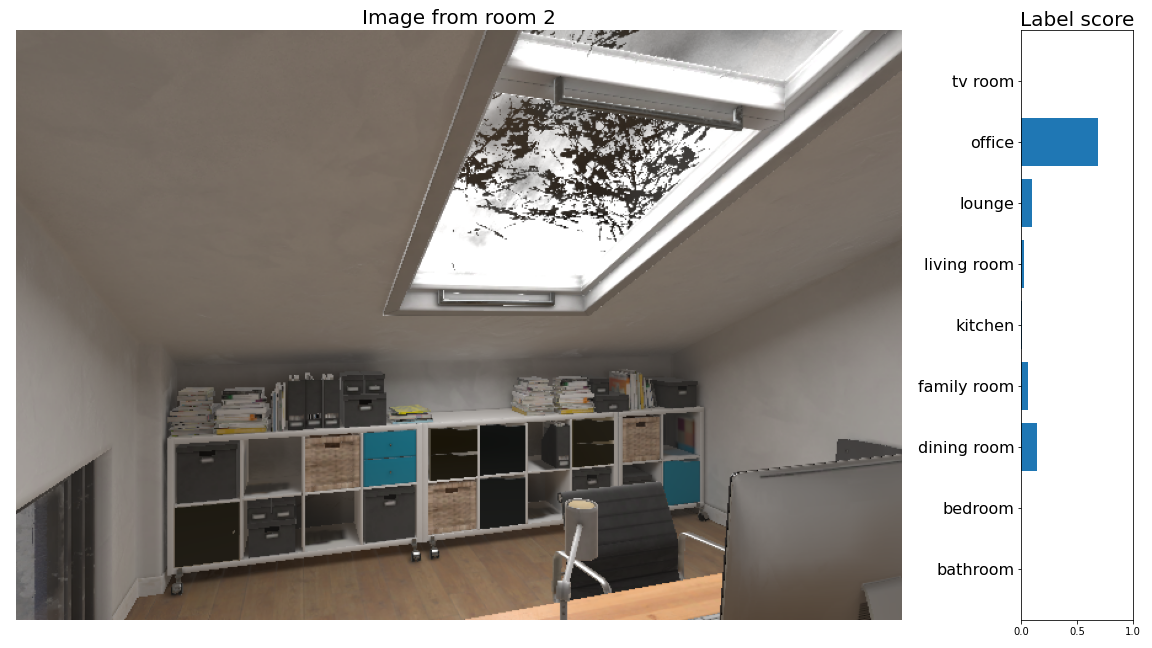}
    \caption{Example visualization of room label predictions on the real scene graph.}
    \label{fig:example_pred}
\end{figure}

\begin{figure*}[ht]
\centering
\subfloat[Rm 0: Dining room. Contains curtains, cabinets, tables, chairs, mirrors, plants, and lighting.]{\includegraphics[width=.3\textwidth]{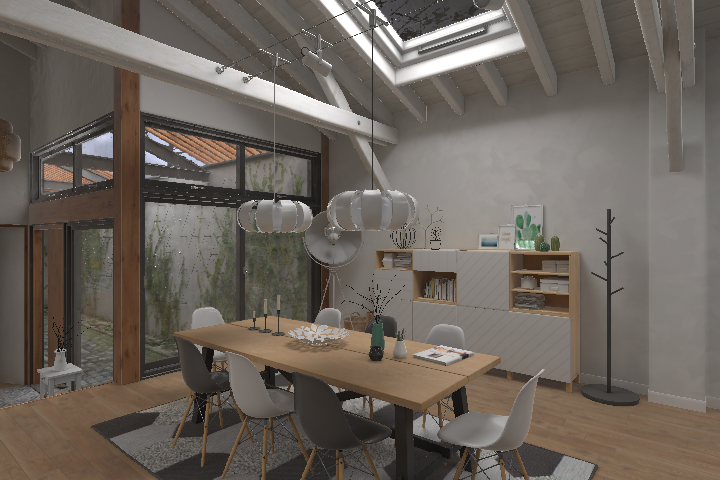}}
\hfill
\subfloat[Rm 1: Bedroom. Contains beds, cabinets, appliances, tables, pictures, mirrors, shelves, lighting, and cushions.]{\includegraphics[width=.3\textwidth]{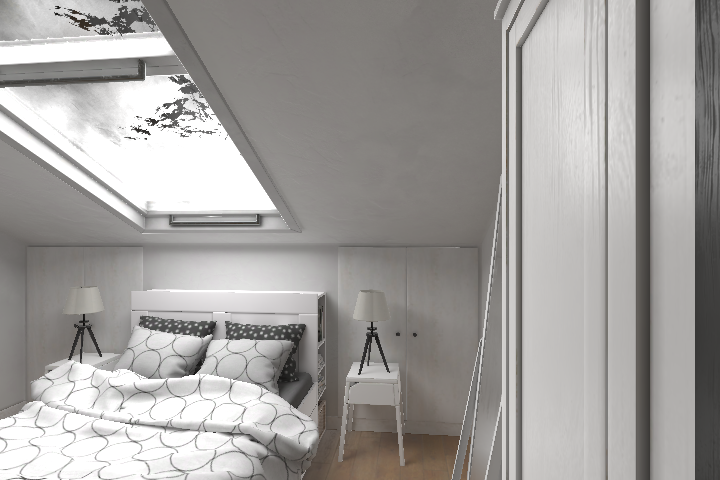}}
\hfill
\subfloat[Rm 2: Office. Contains cabinets, lighting, pictures, and chairs.]{\includegraphics[width=.3\textwidth]{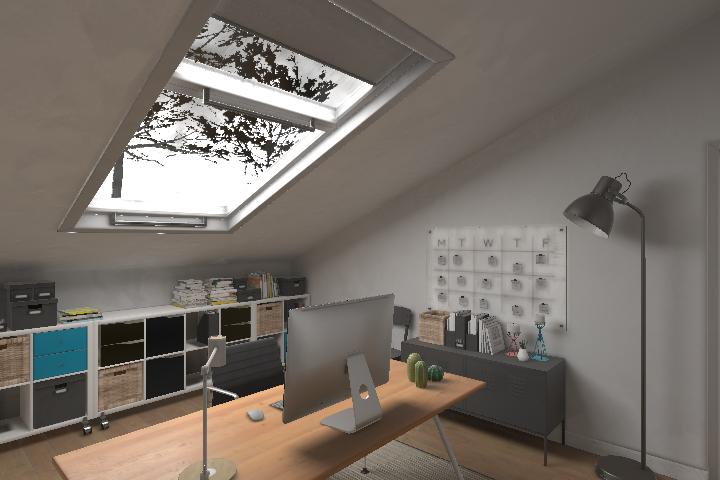}}

\subfloat[Rm 3: Bathroom. Contains towels and toilets.]{\includegraphics[width=.3\textwidth]{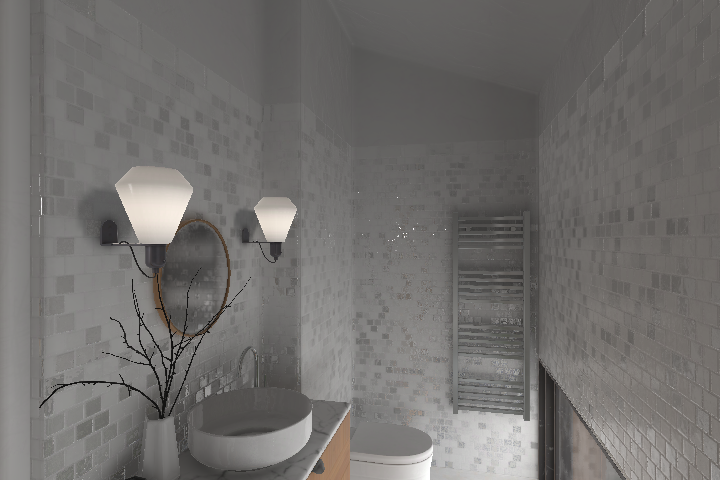}}
\hfill
\subfloat[Rm 4: Living room. Contains curtains, sofas, cabinets, tables, pictures, chairs, shelves, lighting, and cushions.]{\includegraphics[width=.3\textwidth]{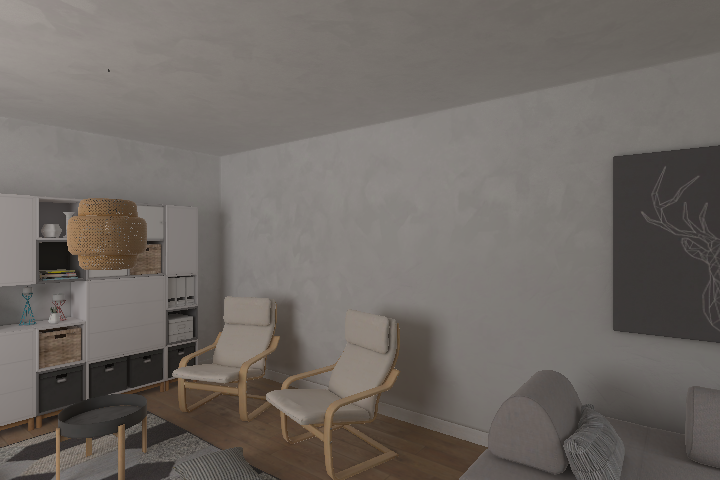}}
\hfill
\subfloat[Rm 5: Kitchen. Contains curtains, cabinets, appliances, tables, pictures, chairs, sinks, shelves, and lighting.]{\includegraphics[width=.3\textwidth]{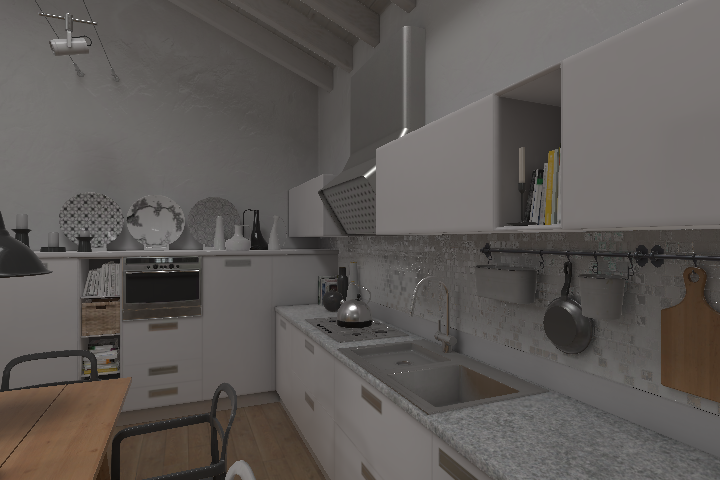}}

\caption{Example egocentric images from each room of the uHumans2 apartment, along with their corresponding ID, ground truth label, and contained objects (as detected by Hydra).}
\label{fig:room_examples}
\end{figure*}

\begin{table*}[ht]
\centering
\resizebox{.8\textwidth}{!}{%
\begin{tabular}{@{}cccccc@{}}
\toprule
\textbf{\begin{tabular}[c]{@{}c@{}}Rm 0\\ Dining Room\end{tabular}} &
  \textbf{\begin{tabular}[c]{@{}c@{}}Rm 1\\ Bedroom\end{tabular}} &
  \textbf{\begin{tabular}[c]{@{}c@{}}Rm 2\\ Office\end{tabular}} &
  \textbf{\begin{tabular}[c]{@{}c@{}}Rm 3\\ Bathroom\end{tabular}} &
  \textbf{\begin{tabular}[c]{@{}c@{}}Rm 4\\ Living Room\end{tabular}} &
  \textbf{\begin{tabular}[c]{@{}c@{}}Rm 5\\ Kitchen\end{tabular}} \\ \midrule
\begin{tabular}[c]{@{}c@{}}Kitchen\\ -10.5\end{tabular} &
  \begin{tabular}[c]{@{}c@{}}Bedroom\\ -5.4\end{tabular} &
  \begin{tabular}[c]{@{}c@{}}Office\\ -16.0\end{tabular} &
  \begin{tabular}[c]{@{}c@{}}Bathroom\\ -1.4\end{tabular} &
  \begin{tabular}[c]{@{}c@{}}Living Room\\ -5.0\end{tabular} &
  \begin{tabular}[c]{@{}c@{}}Kitchen\\ -4.5\end{tabular} \\
\begin{tabular}[c]{@{}c@{}}Dining Room\\ -10.6\end{tabular} &
  \begin{tabular}[c]{@{}c@{}}Family Room\\ -12.0\end{tabular} &
  \begin{tabular}[c]{@{}c@{}}Dining Room\\ -18.6\end{tabular} &
  \begin{tabular}[c]{@{}c@{}}Bedroom\\ -3.4\end{tabular} &
  \begin{tabular}[c]{@{}c@{}}Family Room\\ -9.5\end{tabular} &
  \begin{tabular}[c]{@{}c@{}}Dining Room\\ -10.7\end{tabular} \\
\begin{tabular}[c]{@{}c@{}}Office\\ -16.6\end{tabular} &
  \begin{tabular}[c]{@{}c@{}}Living Room\\ -12.8\end{tabular} &
  \begin{tabular}[c]{@{}c@{}}Lounge\\ -20.9\end{tabular} &
  \begin{tabular}[c]{@{}c@{}}Kitchen\\ -3.6\end{tabular} &
  \begin{tabular}[c]{@{}c@{}}Lounge\\ -9.6\end{tabular} &
  \begin{tabular}[c]{@{}c@{}}Bathroom\\ -11.2\end{tabular} \\ \bottomrule
\end{tabular}%
}
\caption{Top three room-wise logit scores for each room in the uHumans2 apartment, generated by using our fine-tuned VLM approach in conjunction with a scene graph from Hydra. Values are scaled by $1e-3$ for readability. Scores are equivalent to negative loss, so the highest-scoring room is the inferred label. Note that room 0 is the only incorrectly-classified one, and even then, the true label (dining room) is a very close second to the highest-scoring label (kitchen).}
\label{tab:real_dsg_roomwise_logits}
\end{table*}

We present some additional results and analysis for the real scene graph trial. 

Firstly, we showcase some example images drawn from each of the six uHumans2 apartments, including their corresponding ground-truth label in Fig. \ref{fig:room_examples}. We also include an example visualization of the predicted distribution over room labels for a single frame in Fig. \ref{fig:example_pred}. These rooms received the corresponding room-wise label classifications shown in Table \ref{tab:real_dsg_roomwise_logits}, where the top three highest-scoring category per room are noted. In all but one case (room 0), the highest-scoring label is correct. We note that the top categories tend to have overlap of contained objects: e.g., kitchens and bathrooms often both have sinks. 

Moreover, as listed in Fig. \ref{fig:room_examples}, Hydra both uses a small label space (similar to mpcat40) for semantic segmentation and is not entirely reliable in assigning objects to rooms -- there are objects that can be seen in the trajectory video that are not assigned to the rooms they are in, e.g., the shower in the bathroom. 

Nevertheless, the semantic signal it \textit{does} provide already allows our VLM to achieve good room classification performance. We expect better/more informative segmentation and scene graph generation to only improve performance.

\end{document}